\pdfoutput=1
\documentclass[11pt]{article}

\usepackage[]{acl}

\usepackage{times}
\usepackage{latexsym}

\usepackage[T1]{fontenc}
\usepackage[utf8]{inputenc}

\usepackage{microtype}

\usepackage{CJKutf8}

\newcommand{\lx}[1]{[\textcolor{blue}{LX: #1}]}
\newcommand{\bing}[1]{[\textcolor{brown}{\textbf{Bing}: #1}]}

\usepackage{comment}

\usepackage{xcolor}

\usepackage{xcolor,colortbl}
\definecolor{lightblue}{rgb}{0.68, 0.85, 0.9}
\definecolor{lavender}{rgb}{0.9, 0.9, 0.98}
\definecolor{lightyellow}{rgb}{1.0, 1.0, 0.88}
\definecolor{magicmint}{rgb}{0.67, 0.94, 0.82}
\definecolor{palepink}{rgb}{0.98, 0.85, 0.87}
\definecolor{bubbles}{rgb}{0.91, 1.0, 1.0}

\usepackage{graphicx}
\usepackage{tabularx}
\usepackage{multirow}
\usepackage{arydshln}
\usepackage{enumitem}
\usepackage{color}
\usepackage{makecell}
\usepackage{amssymb}
\usepackage{amsthm}

\usepackage{caption}
\usepackage{subcaption}

\usepackage{amsmath}
\DeclareMathOperator*{\argmax}{argmax}

\usepackage{algorithm}
\usepackage{algpseudocode}

\newcommand*{\affaddr}[1]{#1} % No op here. Customize it for different styles.
\newcommand*{\affmark}[1][*]{\textsuperscript{#1}}
\newcommand*{\email}[1]{\text{#1}}

\title{MELM: Data Augmentation with Masked Entity \\Language Modeling for Low-Resource NER}

\author{Ran Zhou\thanks{~~Ran Zhou is under the Joint Ph.D. Program between Alibaba and Nanyang Technological University.}\affmark[~~1,2]\quad Xin Li\thanks{~~Corresponding author}\affmark[~~1] \quad Ruidan He\affmark[1]\quad Lidong Bing\affmark[1]\quad Erik Cambria\affmark[2]\quad Luo Si\affmark[1]\quad Chunyan Miao\affmark[2]\\
	\affaddr{\affmark[1]DAMO Academy, Alibaba Group}\quad
	\affaddr{\affmark[2]Nanyang Technological University, Singapore}\\
	\email{\small{\tt\{ran.zhou, xinting.lx, ruidan.he, l.bing, luo.si\}@alibaba-inc.com}}\\
	\email{\small{\tt\{cambria, ascymiao\}@ntu.edu.sg}} \\
	}

\begin{document}
\maketitle
\begin{abstract}
Data augmentation is an effective solution to data scarcity in low-resource scenarios. However, when applied to token-level tasks such as NER, data augmentation methods often suffer from token-label misalignment, which leads to unsatsifactory performance. In this work, we propose Masked Entity Language Modeling (MELM) as a novel data augmentation framework for low-resource NER. To alleviate the token-label misalignment issue, we explicitly inject NER labels into sentence context, and thus the fine-tuned MELM is able to predict masked entity tokens by explicitly conditioning on their labels. Thereby, MELM generates high-quality augmented data with novel entities, which provides rich entity regularity knowledge and boosts NER performance. When training data from multiple languages are available, we also integrate MELM with code-mixing for further improvement. We demonstrate the effectiveness of MELM on monolingual, cross-lingual and multilingual NER across various low-resource levels. Experimental results show that our MELM presents substantial improvement over the baseline methods.\footnote{Our code is available at \url{https://github.com/RandyZhouRan/MELM/}.}

\end{abstract}

\section{Introduction}
%\bing{for the whole paper, insert reference papers at appropriate places. no need to discuss specific details, say when you are mentioning some commonsense or conclusion, put some references behind.}
Named entity recognition (NER) is a fundamental NLP task which aims to locate named entity mentions and classify them into predefined categories. As a subtask of information extraction, it serves as a key building block for information retrieval~\citep{banerjee2019information}, question answering~\citep{fabbri2020template} and text summarization systems~\citep{nallapati2016abstractive} etc. However, except a few high-resource languages / domains, the majority of languages / domains have limited amount of labeled data. 

Since manually annotating sufficient labeled data for each language / domain is expensive, low-resource NER~\citep{cotterell2017low, feng2018improving, zhou2019dual, rijhwani2020soft} has received increasing attention in the research community over the past years.
%As an efficient solution to data scarcity in low-resource scenarios, data augmentation has been widely adopted in CV and speech tasks, where simple techniques like \lx{==> such as} rotation, cropping and distortion produces\lx{produce==>are developed to produce} diversified augmented data. However, data augmentation for NLP is more difficult due to the discreteness of human languages. \lx{++ For example, }Randomly replacing a word might change the semantics of a sentence, resulting in noisy samples in the training data. As a result, many works on data augmentation for NLP focuses on sentence level tasks~\citep{wei2019eda, kobayashi2018contextual, wu2019conditional, kumar2020data}, which are less sensitive towards aforementioned noise from token-level disturbance. They generate modifications or paraphrases of the original sentence while keeping the sentence-level label unchanged. In contrast, on token-level tasks such as NER, the alignment between tokens and labels need to be handled carefully. For example, simple word replacement methods for NER could replace an entity token with alternatives that mismatch the original label, leading to incompatible token and label sequences\lx{The origanization of this paragraph is poor!}. 
As an effective solution to data scarcity in low-resource scenarios, data augmentation enlarges the training set by applying label-preserving transformations. Typical data augmentation methods for NLP include (1) word-level modification~\citep{wei2019eda, kobayashi2018contextual, wu2019conditional, kumar2020data} and (2) back-translation~\citep{sennrich2015improving,fadaee2017data,dong2017learning,yu2018fast}. 
%Despite their effectiveness on sentence-level tasks, it is difficult to apply them to token-level tasks such as NER. Word-level modification might replace an entity with alternatives that mismatch the original entity label, leading to token-label misalignment. On the other hand, back-translation on NER requires careful label alignment due to word order changes between different languages. 

Despite the effectiveness on sentence-level tasks, they suffer from the token-label misalignment issue when applied to token-level tasks like NER. More specifically, word-level modification might replace an entity with alternatives that mismatch the original label. 
Back-translation creates augmented texts that largely preserve the original content. However, it hinges on external word alignment tools for propagating the labels from the original input to the augmented text, which has proved to be error-prone.

%\lx{THE THIRD PARAGRAPH==>To achieve data augmentation for token-level tasks,~\citet{dai2020analysis} proposed to randomly substitute entity mentions with existing entities of the same class. It avoided the token-label misalignment issue but the entity diversity does not change. Besides, the substituted entity might not fit into the original context.~\citet{li2020conditional} conquered the token-label misalignment issue by increasing the context diversity only, where they replaced context (‘O’) tokens using MASS~\citep{song2019mass} and left the entities (aspect terms) completely unchanged. However, according to the NER evaluations in~\citet{lin2020rigorous}, the context augmentation gave marginal improvement on pre-trained language model. Our preliminary results on low-resource NER (see Figure~\ref{fig:evc}) also demonstrate that diversifying entities in the training data is more effective than introducing more context patterns. Specifically, when given 100 gold samples, ``Add Entity'' produces 100 augmentation samples by borrowing name entities from other gold samples and inserting them into the context of the given 100 gold samples. While ``Add Context'' produces 100 augmentation samples by inserting the entities of the given 100 gold samples into the context of another 100 gold samples. ADD CONCLUSION HERE}

%\zr{THE THIRD PARAGRAPH==>
To apply data augmentation on token-level tasks,~\citet{dai2020analysis} proposed to randomly substitute entity mentions with existing entities of the same class. They avoided the token-label misalignment issue but the entity diversity does not increase. Besides, the substituted entity might not fit into the original context.~\citet{li2020conditional} avoided the token-label misalignment issue by only diversifying the context, where they replaced context (having ‘O’ label) tokens using MASS~\citep{song2019mass} and left the entities (i.e. aspect terms in their task) completely unchanged. However, according to the NER evaluations in~\citet{lin2020rigorous}, augmentation on context gave marginal improvement on pretrained-LM-based NER models.

$ $

\begin{figure}[t]
    \centering
    \includegraphics[width=1\columnwidth]{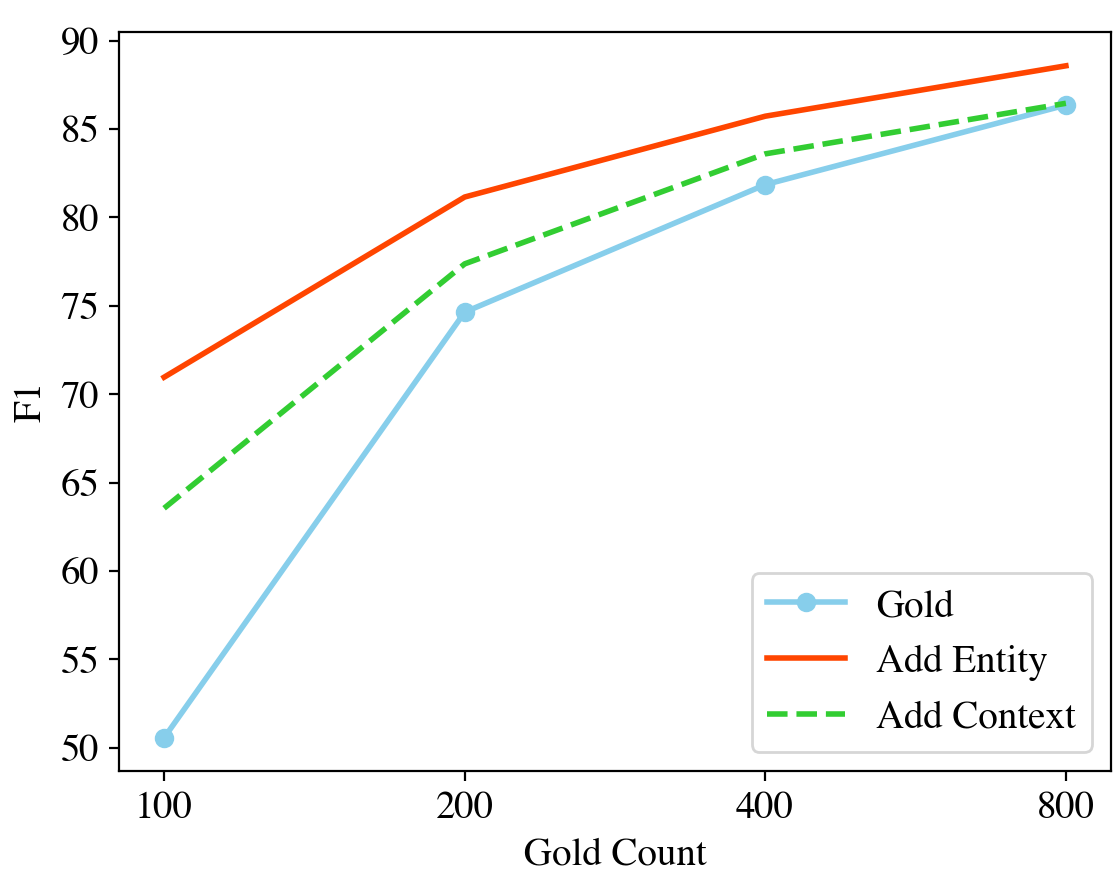}
    \caption{Effectiveness comparison between diversifying entities and diversifying context. Given $N$ gold samples, \textbf{Add Entity} substitutes their entities with new entities from extra gold samples. In contrary, \textbf{Add Context} reuses existing entities and inserts them into context of extra gold samples. Both methods yield $N$ augmented samples.}
    \label{fig:evc}
\end{figure}

\begin{figure*}
    \centering
    \begin{subfigure}[b]{\textwidth}
         \centering
         \includegraphics[width=\textwidth]{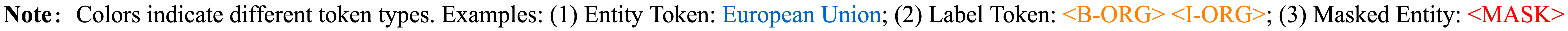}
         \label{fig:note}
    \end{subfigure}
     %\hfill
     \begin{subfigure}[b]{0.23\textwidth}
        \centering
        \includegraphics[width=\textwidth]{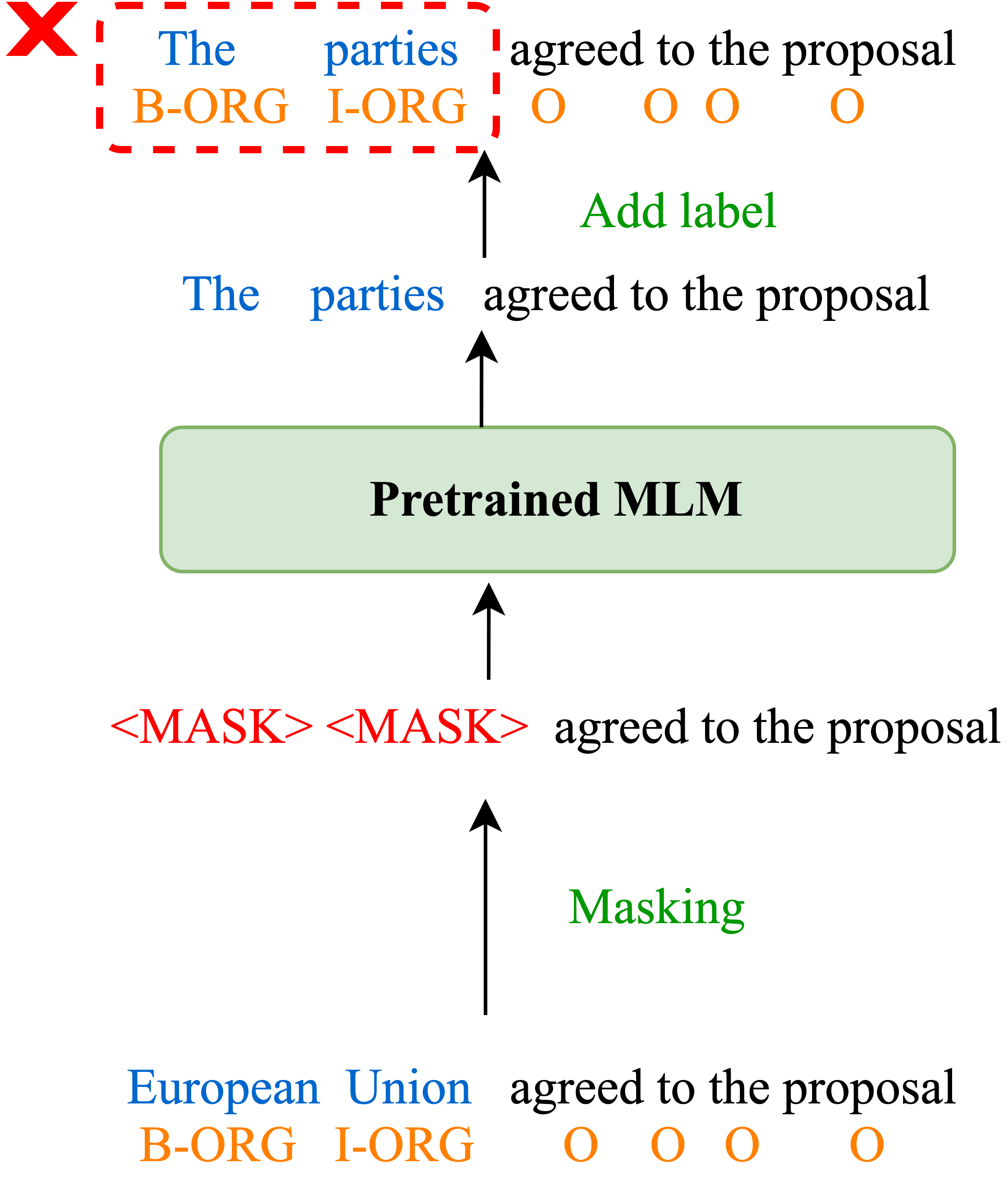}
        \caption{}
        \label{fig:MELM_0}
    \end{subfigure}
    %\hfill
    \hspace{0.02\textwidth}
    \begin{subfigure}[b]{0.22\textwidth}
        \centering
        \includegraphics[width=\textwidth]{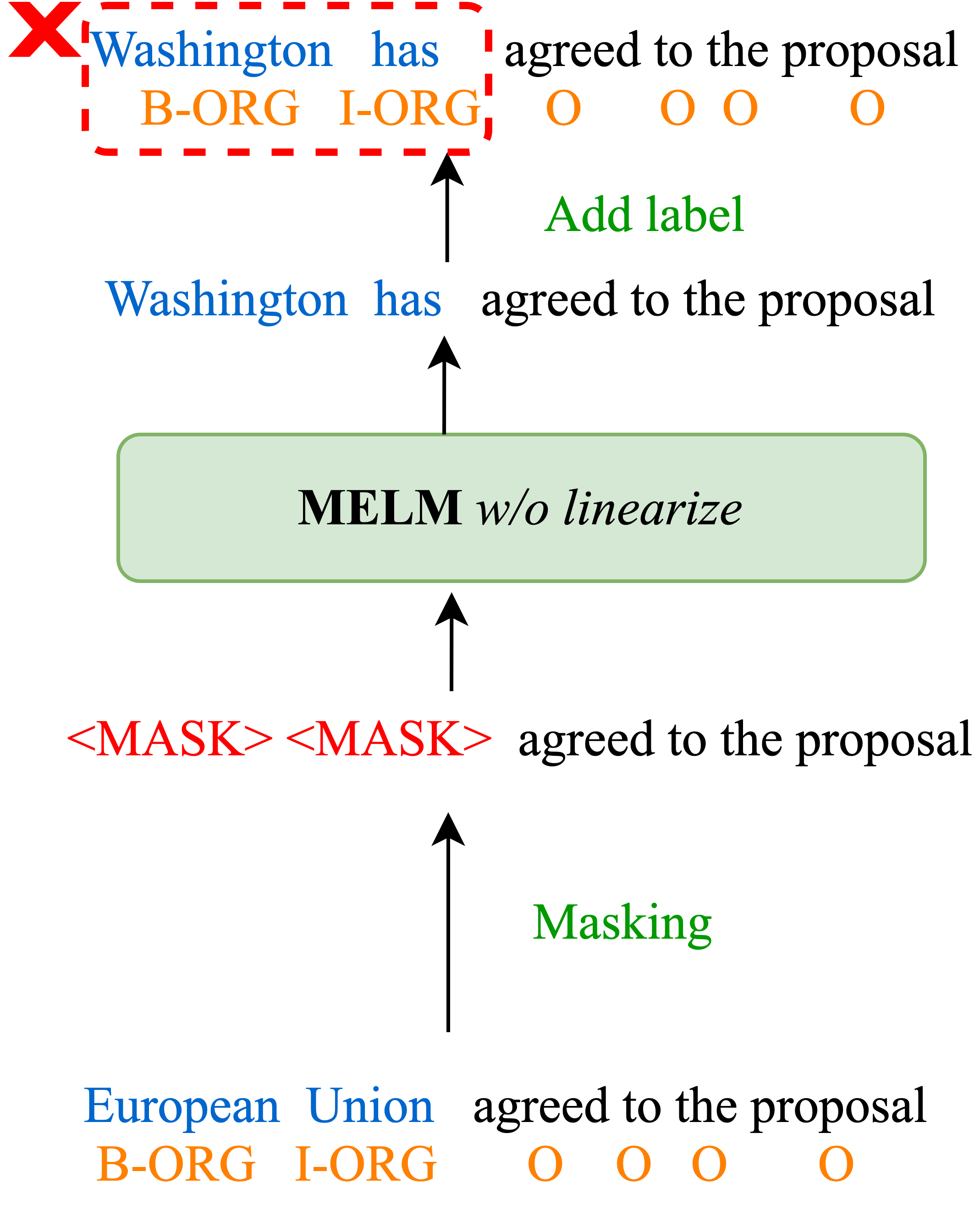}
        \caption{}
        \label{fig:MELM_a}
    \end{subfigure}
    %\hfill
    \hspace{0.02\textwidth}
    \begin{subfigure}[b]{0.43\textwidth}
        \centering
        \includegraphics[width=\textwidth]{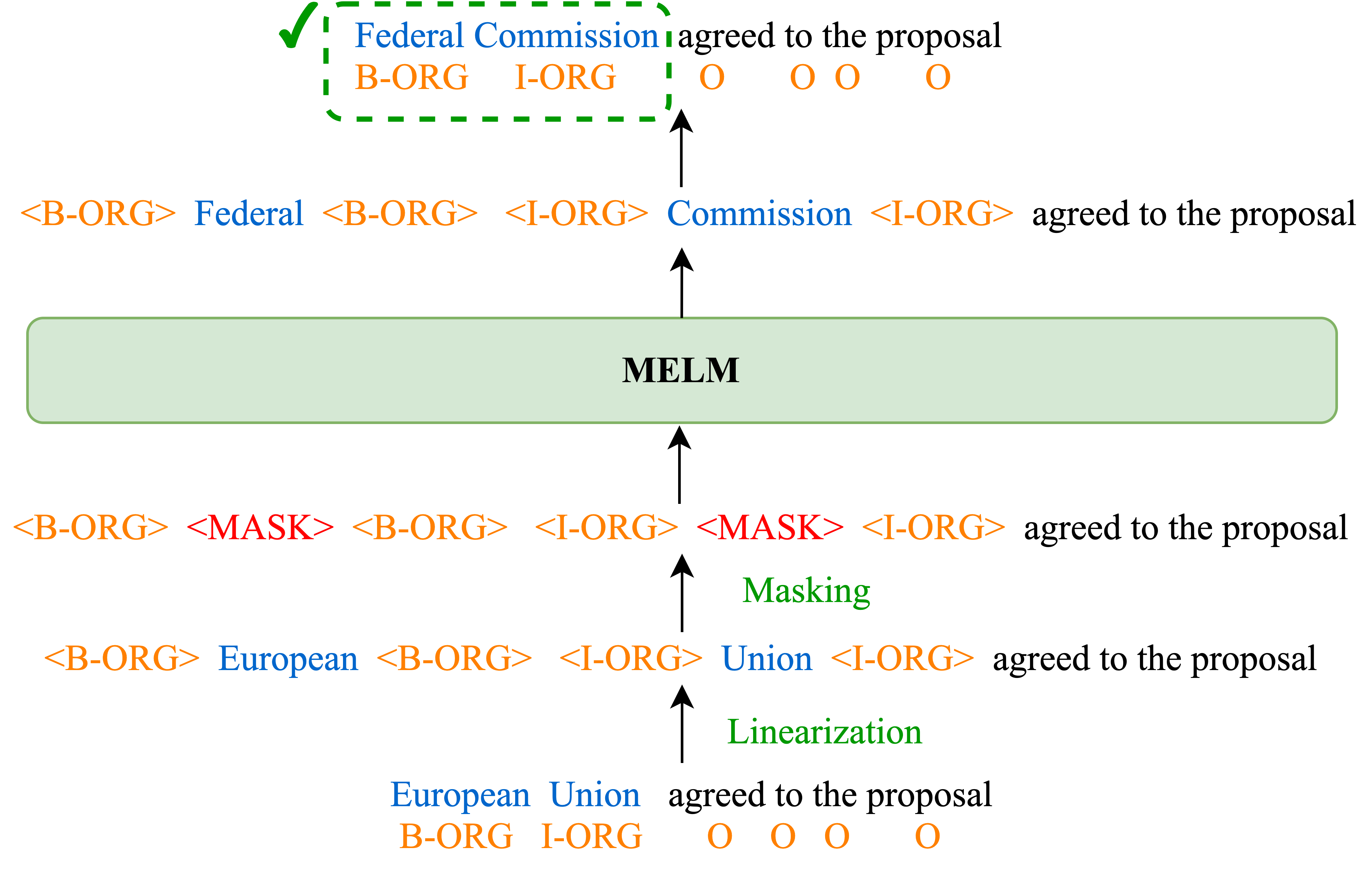}
        \caption{}
        \label{fig:MELM_b}
    \end{subfigure}
    % \hfill
    % \begin{subfigure}[b]{0.375\textwidth}
    %     \centering
    %     \includegraphics[width=\textwidth]{MELM_Multi.png}
    %     \caption{}
    %     \label{fig:MELM_c}
    % \end{subfigure}    
    %\includegraphics[width=16cm]{MELM_CM}
    \caption{Comparison of different data augmentation methods, color printing is preferred. (a) augmentation with pretrained MLM (b) augmentation with MELM without linearization (c) augmentation with MELM} 
    %(c) augmentation with MELM on code-mixed data \zr{change (a) to MELM without linearization? change example to multi-word entity?}}
    \label{fig:MELM}
\end{figure*}

Our preliminary results on low-resource NER (see Figure~\ref{fig:evc}) also demonstrate that diversifying entities in the training data is more effective than introducing more context patterns.
Inspired by the aforementioned observations, we propose Masked Entity Language Modeling (MELM) as a data augmentation framework for low-resource NER, which generates augmented data with diverse entities while alleviating the challenge of token-label misalignment. MELM is built upon pretrained Masked Language Models (MLM), and it is further fine-tuned on corrupted training sentences with only entity tokens being randomly masked to facilitate entity-oriented token replacement. 
%MELM then generates augmented data by masking and replacing entity tokens in the original training sentences\lx{this sentence can be removed?}.  
Simply masking and replacing entity tokens using the finetuned MLM is still insufficient because the predicted entity might not align with the original label.
Taking the sentence shown in Figure~\ref{fig:MELM_a} as an example, after masking the named entity ``European Union'' (Organization), the finetuned MLM could predict it as ``Washington has''. Such prediction fits the context but it is not aligned with the original labels.
To alleviate the misalignment, our MELM additionally introduces a labeled sequence linearization strategy, which respectively inserts one label token before and after each entity token and regards the inserted label tokens as the normal context tokens during masked language modeling. 
Therefore, the prediction of the masked token is conditioned on not only the context but the entity's label as well. 
%Then the fine-tuned MLM can utilize the rich knowledge from pretrained MLM, such that the generated augmentation sentences present diverse entities with compatible labels that fit into the context as well. 

After injecting label information and finetuning on the label-enhanced NER data, our MELM can exploit rich knowledge from pre-training to increase entity diversity while greatly reducing token-label misalignment.
Code-mixing~\citep{singh2019xlda,qincosda, zhang-etal-2021-cross} achieved promising results by creating additional code-mixed samples using the available multilingual training sets, which is particularly beneficial when the training data of each language is scarce. Fortunately, in the scenarios of multilingual low-resource NER, our MELM can also be applied on the code-mixed examples for further performance gains.  We first apply code-mixing by replacing entities in a source language sentence with the same type entities of a foreign language. However, even though token-label alignment is guaranteed by replacing with entities of the same type, the candidate entity might not best fit into the original context (for example, replacing a government department with a football club). To solve this problem, we propose an entity similarity search algorithm based on bilingual embedding to retrieve the most semantically similar entity from the training entities in other languages. Finally, after adding language markers to the code-mixed data, we use them to fine-tune MELM for generating more code-mixed augmented data.

To summarize, the main contributions of this paper are as follows: (1) we present a novel framework which jointly exploits sentence context and entity labels for entity-based data augmentation. It consistently achieves substantial improvement when evaluated on monolingual, cross-lingual, and multilingual low-resource NER; (2) the proposed labeled sequence linearization strategy effectively alleviates the problem of token-label misalignment during augmentation; (3) an entity similarity search algorithm is developed to better bridge entity-based data augmentation and code-mixing in multilingual scenarios.

\section{Method}\label{sec: method}

%As demonstrated in~\citet{lin2020rigorous} and our experiments in Figure~\ref{fig:evc}, pretrained-LM-based NER models can easily capture context pattern given a small amount of data. In contrast, increasing the diversity of entities provides a larger boost on NER performance through richer name regularity knowledge. 
%Therefore, we aim to infuse diversified entities into the training set, by masking original entities and generating novel substitutes using a Masked Language Model (MLM). 

%However, simply adopting a pretrained MLM will bring in the token-label misalignment problem, which is critical for NER\lx{==> simply adopting MLM for data augmentation is unsatisfactory for NER because it cannot guarantee the compatibility between the generated token and the original label}. As the entity type information is not visible for the pretrained MLM, its prediction is purely conditioned on the context words and the generated entity might not align with the original label. Taking the masked sentence ``\(\langle\)mask\(\rangle\) launches the new iPhone'' derived from a training sentence ``Cook launches the new iPhone'' (see Figure~\ref{fig:MELM_a}) as an example, "Cook" is labeled with PER but the masked token could be predicted as an organization (ORG) "Apple".  
%We propose MELM as a data augmentation framework for low-resource NER. We first perform labeled sequence linearization which injects label information into the input sequence, to help the MLM condition its prediction on the correct entity type (Section~\ref{ssec:linearize})
Fig.~\ref{fig:MELM_b} presents the work flow of our proposed data augmentation framework. We first perform labeled sequence linearization to insert the entity label tokens into the NER training sentences (Section~\ref{ssec:linearize}).
Then, we fine-tune the proposed MELM on linearized sequences (Section~\ref{ssec:fine-tune}) and create augmented data by generating diverse entities via masked entity prediction (Section~\ref{ssec:gen_aug}). 

The augmented data undergoes post-processing (Section~\ref{ssec:postp}) and is combined with the original training set for training the NER model. Algorithm~\ref{alg:MELM} gives the pseudo-code for the overall framework.
%Under multilingual settings, we further leverage code-mixing by proposing a phrase similarity search algorithm (Section~\ref{ssec:code-mix}). The generated code-mixed data can be populated with the a similar fine-tuning and generation procedure as well
Under multilingual scenarios, we propose an entity similarity search algorithm as a refined code-mixing strategy (Section~\ref{ssec:code-mix}) and apply our MELM on the union set of gold training data and code-mixed data for further performance improvement.

\begin{algorithm*}[h!]
\small
\caption{Masked Entity Language Modeling (MELM)}\label{alg:MELM}
\begin{algorithmic}
\State \textbf{Given} $\mathbb{D_{\text{train}}}, \mathcal{M}$ \Comment{Given gold traning set $\mathbb{D_{\text{train}}}$ and pretrained MLM $\mathcal{M}$}
\hspace{\algorithmicindent}\State $\mathbb{D}_{\text{masked}} \gets \emptyset, \mathbb{D}_{\text{aug}} \gets \emptyset$
\For{$\{X, Y\} \in \mathbb{D_{\text{train}}}$}
    \State $\tilde{X} \gets \textsc{Linearize}(X, Y)$ \Comment{Labeled sequence linearization}
    \State $\tilde{X} \gets \textsc{FinetuneMask}(\tilde{X}, \eta)$ \Comment{Randomly mask entities for fine-tuning}
    \State $\mathbb{D}_{\text{masked}} \gets \mathbb{D}_{\text{masked}} \cup \{\tilde{X}\}$ 
\EndFor
\State $\mathcal{M}_{\text{finetune}} \gets \textsc{Finetune}(\mathcal{M}, \mathbb{D}_{\text{masked}})$ \Comment{Fine-tune MELM on masked linearized sequences}
\For{$\{X, Y\} \in \mathbb{D}_{\text{masked}}$}
    \State \textbf{repeat} $R$ \textbf{times}:
        \State \hskip1.5em $\tilde{X} \gets \textsc{Linearize}(X, Y)$ \Comment{Labeled sequence linearization}
        \State \hskip1.5em $\tilde{X} \gets \textsc{GenMask}(\tilde{X}, \mu)$ \Comment{Randomly mask entities for generation}
        \State \hskip1.5em $X_{\text{aug}} \gets \textsc{RandChoice}(\mathcal{M}_{\text{finetune}}(\tilde X), \text{Top } k=5)$ \Comment{Generate augmented data with fine-tuned MELM}
        \State \hskip1.5em $\mathbb{D}_{\text{aug}} \gets \mathbb{D}_{\text{aug}} \cup \{X_{\text{aug}}\}$
\EndFor
\State $\mathbb{D}_{\text{aug}} \gets \textsc{PostProcess} (\mathbb{D}_{\text{aug}})$ \Comment{Post-processing}
\State \textbf{return} $\mathbb{D_{\text{train}}} \cup \mathbb{D}_{\text{aug}}$
\end{algorithmic}
\end{algorithm*}

\subsection{Labeled Sequence Linearization}
\label{ssec:linearize}

% To enable the MLM to condition its prediction on the entity type, we first inject label information into the input sequence via labeled sequence linearization. 
% Specifically, as shown in Figure~\ref {fig:MELM_b}, an entity token’s label is inserted before and after the entity, which is treated equally as normal context tokens by the MLM. The inserted label tokens are added to the vocabulary as special tokens. 
% During fine-tuning, the model utilizes the label information from added tokens to capture the dependency between an entity type and its corresponding entity cluster. To accelerate the training of added label tokens, we initialize their embeddings with semantically related seed words (e.g. ``person’’ for $\langle$ B-PER $\rangle$). 
% As a result, the linearized sequence resembles a natural sentence and the MLM is able to exploit the pretrained semantic information of the seed words to facilitate prediction of the masked tokens.
To minimize the amount of generated tokens incompatible with the original labels, we design a labeled sequence linearization strategy to explicitly take label information into consideration during masked language modeling. Specifically, as shown in Figure~\ref {fig:MELM_b}, we add the label token before and after each entity token and treat them as normal context tokens. The yielded linearized sequence is utilized to further finetune our MELM so that its prediction is additionally conditioned on the inserted label tokens. Note that, we initialize the embeddings of label tokens with those of tokens semantically related to the label names (e.g., “organization” for $\langle$ B-ORG $\rangle$). By doing so, the linearized sequence is semantically closer to a natural sentence and the difficulty of finetuning on linearized sequence could be reduced~\cite{kumar2020data}.

$ $

\subsection{Fine-tuning MELM}
\label{ssec:fine-tune}

%We then finetune a pretrained MLM on the linearized training samples\lx{This sentence can be deleted?}. 
Unlike MLM, only entity tokens are masked during MELM fine-tuning. At the beginning of each fine-tuning epoch, we randomly mask entity tokens in the linearized sentence $X$ with masking ratio $\eta$. 

Then, given the corrupted sentence $\tilde{X}$ as input, our MELM is trained to maximize the probabilities of the masked entity tokens and reconstruct the linearized sequence $X$:
% \begin{equation}
% \label{eqn:L_MLM}
% \small
% \mathrm{L_{MLM}=-\frac{1}{n}\sum_{i=1}^{n}\sum_{j=1}^{\left | V \right |}y_{ij}log(\hat{y}_{ij}~|~\mathbf{\tilde{x})}}
% \end{equation}
%where $|V|$ is the size of vocabulary and $n$ is the number of tokens. 
\begin{equation}
    \max_{\theta }~~~\mathrm{log}~p_{\theta }(X|\tilde{X}) \approx \sum_{i=1}^{n}~m_{i}~\mathrm{log}~p_{\theta }(x_{i}|\tilde{X})
\end{equation}
where $\theta$ represents the parameters of MELM, $n$ is the number of tokens in $\tilde{X}$, $x_{i}$ is the original token in $X$, $m_{i}=1$ if $x_{i}$ is masked and otherwise $m_{i}=0$.
Through the above fine-tuning process, the proposed MELM learns to make use of both contexts and label information to predict the masked entity tokens. 
As we will demonstrate in Section~\ref{ssec: analysis_align}, the predictions generated by the fine-tuned MELM are significantly more coherent with the original entity label, compared to those from other methods.

\subsection{Data Generation}
\label{ssec:gen_aug}

% With the fine-tuned MELM, we proceed to generate augmented data from the original training set. 
% Given a linearized sequence with masked entity tokens, we generate the top 5 predictions on each masked token and randomly select one from them as the substitution to\lx{==> randomly sample one from } the original entity token. Note that we do not simply adopt the top prediction to avoid memorization and unchanged augmented data, as we fine-tune and augment on the same training dataset. 

% \zr{Given a linearized ... ==> For a masked token in the corrupted linearized sequence\lx{==> At each step of prediction/generation?}, MELM outputs the probability of each word\lx{==> token} in the vocabulary being the ground truth token\lx{++ (i.e., the masked entity token)}. However, as we fine-tune and generate on the same dataset, MELM is likely to memorize the ground truth token, and simply picking the top prediction will result in an augmented sample identical to the original training sample. Therefore, we adopt top-k sampling strategy~\citep{fan2018hierarchical} to randomly sample from the top-5 predictions as the substitution to the masked token. Thereby, while we ensure the prediction fit into the context and align with the entity label, the diversity of augmented data is also greatly enhanced.}

To generate augmented training data for NER, we apply the fine-tuned MELM to replace entities in the original training samples. Specifically, given a corrupted sequence, MELM outputs the probability of each token in the vocabulary being the masked entity token. However, as the MELM is fine-tuned on the same training set, directly picking the most probable token as the replacement is likely to return the masked entity token in the original training sample, and might fail to produce a novel augmented sentence. Therefore, we propose to \textit{randomly sample} the replacement from the top \textit{k} most probable components of the probability distribution. 
Formally, given the probability distribution $P(x_i|\tilde{X})$ for a masked token, we first select a set $V^k_i \subseteq V$ of the $k$ most likely  candidates. Then, we fetch the replacement $\hat{x}_i$ via random sampling from $V^k_i$.
%\lx{==> To generate augmented training data for NER, we apply the fine-tuned MELM on the original training set again. Specifically, at each generation step, MELM outputs the probability of each token in vocabulary being the ground truth (i.e., the masked entity token). However, as the MELM is fine-tuned on the same training set, directly picking the most probable token as the replacement is very likely to fail in producing a novel sentence because the token with the highest probability is the masked entity token itself in most cases. In this paper, we propose to \textit{randomly sample} the replaced entity token from the top \textit{k} most probable components of the probability distribution. Formally, given the probability distribution $P(x_i|\tilde{X})$ at the $i$-th timestep, we first select a set $V^k_i \subseteq V$ of the $k$ most likely  candidates. Then, we fetch the replaced token $\hat{x}_i$ via random sampling from $V^k_i$.} 
% After removing the inserted label tokens, we use the generated sequence as an augmented example\lx{==> 
After obtaining the generated sequence, we remove the label tokens and use the remaining parts as the augmented training data. For each sentence in the original training set, we repeat the above generation procedure $R$ rounds to produce $R$ augmented examples. %$R$ is a hyperparameter\lx{Remove this sentence}.

% To increase the diversity of augmented data, we adopt a different masking strategy from train time. For an entity mention comprising of \(n\) tokens, we first randomly sample a total number of $m$ tokens to be masked, where $m$ is a number drawn from Gaussian distribution \(m \sim \mathcal{N}(\mu=\xi n,\,\sigma^{2})\) in range $[1,n]$. \(\xi\) is a coefficient controlling the Gaussian mean as a proportion of the entity's length, and $\sigma^{2}$ is Gaussian variance. We set $\sigma=1$ and treat $\xi$ as a hyperparameter. This ensures that single token entities will always be masked for augmentation. Meanwhile, the same sentence will have different augmentation results from different rounds of augmentation, resulting in more varied augmented data.

To increase the diversity of augmented data, we adopt a different masking strategy from train time.
% \lx{==> we adopt a different masking strategy during data generation}. For an entity mention comprising of \(n\) tokens, we first randomly sample a total number of $m$ tokens to be masked:
For each entity mention comprising of $n$ tokens, we randomly sample a dynamic masking rate $\epsilon$ from Gaussian distribution $\mathcal{N}(\mathrm{\mu,\,\sigma^{2}})$, where the Gaussian variance $\sigma^{2}$ is set as $1/n^{2}$.
% \begin{equation}
%     %\small
%     m \sim \mathcal{N}(\mathrm{\mu=\xi n,\,\sigma^{2}=1})~~~\textbf{s.t.}~m \in \mathrm{[1,n]}
% \end{equation}
% where $\mathcal{N}$ is the Gaussian distribution, \(\xi\) is a hyperparameter controlling the Gaussian mean as a proportion of the entity's length. 
Thus, the same sentence will have different masking results in each of the $R$ augmentation rounds, resulting in more varied augmented data.

\subsection{Post-Processing}
\label{ssec:postp}

To remove noisy and less informative samples from the augmented data, the generated augmented data undergoes post-processing. 
%1) We discard short sentences with less than 10 tokens. 
%As these sentences contain limited context, the predicted entities are less reliable and could introduce noise for training the NER model\lx{``As these sentences...'' is this sentence necessary?}; 
%2) A baseline NER model trained with only the original gold data is used to assign NER tags to the augmented sentences. We only keep the augmented sentences whose predicted labels are consistent with their original labels\lx{==> 
Specifically, we train a NER model with the available gold training samples and use it to automatically assign NER tags to each augmented sentence. Only augmented sentences whose predicted labels are consistent with the their original labels are kept.
The post-processed augmented training set $\mathbb{D}_{\text{aug}}$ is combined with the gold training set $\mathbb{D}_{\text{train}}$ to train the final NER tagger.
%Note that we also remove sentences with only O ("Other") tag in the label sequence before our MELM fine-tuning since they do not contain any named entities to be augmented.

\subsection{Extending to Multilingual Scenarios}
\label{ssec:code-mix}
%where the golden training sets $\mathbb{D}^{\ell}_{\text{gold}}$ ($\ell \in \mathbb{L}$) are available for a set $\mathbb{L}$ of several languages
% On multilingual low-resource NER, it is straightforward to separately apply the proposed MELM on language-specific data. However, we further attempt to enable MELM on top of the code-mixing technique, which proved to be effective in enhancing multilingual learning~\citep{singh2019xlda, qincosda, zhang-etal-2021-cross}. In this paper, with the aim of bridging MELM augmentation and code-mixing, we propose a phrase similarity search algorithm to perform MELM-friendly code-mixing. 

When extending low-resource NER to multilingual scenarios, it is straightforward to separately apply the proposed MELM on language-specific data for performance improvement. Nevertheless, it offers higher potential to enable MELM on top of code-mixing techniques, which proved to be effective in enhancing multilingual learning~\citep{singh2019xlda, qincosda, zhang-etal-2021-cross}. In this paper, with the aim of bridging MELM augmentation and code-mixing, we propose an entity similarity search algorithm to perform MELM-friendly code-mixing. 

\begin{figure}[h]
    \centering
    \includegraphics[width=0.99\columnwidth]{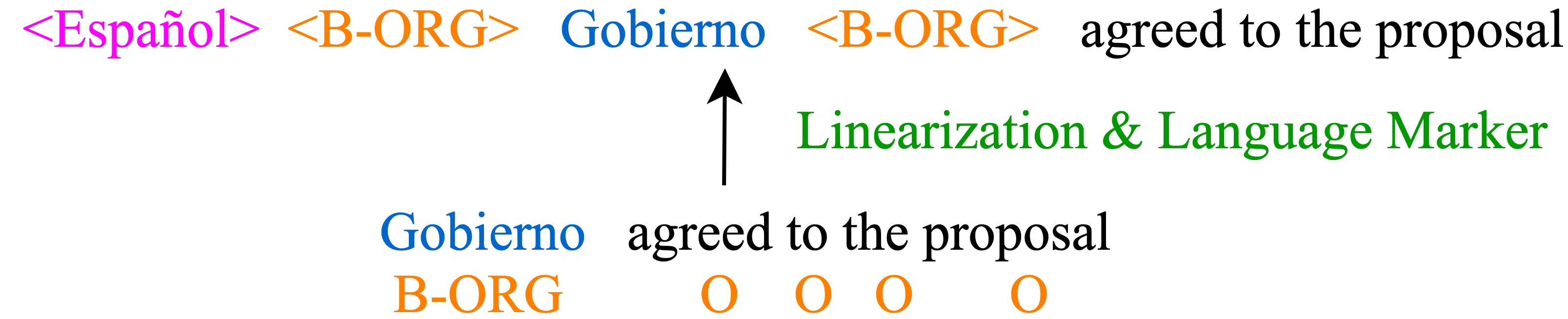}
    \caption{Applying MELM on gold and code-mixed data. Language markers (e.g.,~\textcolor{magenta}{<Español>}) are inserted during linearization.}
    \label{fig:MELM_multi}
\end{figure}

Specifically, given the gold training sets $\{\mathbb{D}^{\ell}_{\text{train}}~|~\ell \in \mathbb{L}\}$ over a set $\mathbb{L}$ of languages, we first collect label-wise entity sets $\mathbb{E}^{\ell, y}$, which consists of the entities appearing in $\mathbb{D}^{\ell}_{\text{train}}$ and belonging to class $y$. To apply code-mixing on a source language sentence $X^{\ell_{\text{src}}}$, we aim to substitute
a mentioned entity $\mathbf{e}$ of label $y$ with a target language entity $\mathbf{e}_{sub} \in \mathbb{E}^{\ell_{\text{tgt}}, y}$, where the target language is sampled as $\ell_\text{tgt} \sim \mathcal{U}(\mathbb{L} \setminus \{\ell_{\text{src}}\})$. 
%Instead of randomly selecting $e_{sub}$ from $\mathbb{E}^{\ell_{\text{tgt}}, y}$, we focus on the phrase similarity between $e$ and $e_{sub}$ to improve the semantic fluency of generated code-mixed data.
%Concretely, we calculate the representation of a multi-token entity as the average pooling of the MUSE bilingual embeddings~\citep{conneau2017word} of its component tokens\lx{``we focus on phrase similarity search...''==> 
Instead of randomly selecting $\mathbf{e}_{sub}$ from $\mathbb{E}^{\ell_{\text{tgt}}, y}$, we choose to retrieve the entity with the highest semantic similarity to $\mathbf{e}$ as $\mathbf{e}_{sub}$. Practically, we introduce MUSE bilingual embeddings~\citep{conneau2017word} and calculate the entity's embedding $\texttt{Emb}(\mathbf{e})$ by averaging the embeddings of the entity tokens:
% \begin{equation}
% \label{eqn:muse_emb}
% \small
% \mathbf{Emb}(e) = \textsc{AvgPool}(\left \{\textsc{MUSE}_{\ell_{\text{src}},\ell_{\text{tgt}}}(e_{i})~|~e_{i}\in e \right \})
% \end{equation}
\begin{equation}
    \texttt{Emb}(\mathbf{e}) = \frac{1}{|\mathbf{e}|} \sum^{|\mathbf{e}|}_{i=1} \textsc{MUSE}_{\ell_{\text{src}},\ell_{\text{tgt}}}(\mathbf{e}_i) 
\end{equation}
where $\textsc{MUSE}_{\ell_{\text{src}},\ell_{\text{tgt}}}$ denotes the $\ell_{\text{src}}-\ell_{\text{tgt}}$ aligned embeddings and $\mathbf{e}_i$ is the $i$-th token of $\mathbf{e}$. Next, we obtain the target-language entity $\mathbf{e}_{sub}$ semantically closest to $\mathbf{e}$ as follows:
\begin{equation}
    \mathbf{e}_{sub}= \argmax_{\tilde{\mathbf{e}} \in \mathbb{E}^{\ell_\text{tgt}, y}}~f(\texttt{Emb}(\mathbf{e}), \texttt{Emb}(\tilde{\mathbf{e}}))
\end{equation}
$f(\cdot, \cdot)$ here is the cosine similarity function. The output entity $\mathbf{e}_{sub}$ is then used to replace $\mathbf{e}$ to create a code-mixed sentence more suitable for MELM augmentation. 
%\lx{THE PARAGRAPH BELOW ==> Regarding the augmented data generation, since the code-mixed data contains entities from multiple languages, we also prepend a language marker to the entity token to help MELM differentiate different languages, as shown in Figure~\ref{fig:MELM_multi}.}
% After ranking the cosine similarity between $e$ and candidate entities $\tilde e \in \mathbb{E}^{\ell_\text{tgt}, y}$, the most similar entity is used to replace the original entity and obtain a code-mixed sentence: 
% \begin{equation}
% \label{eqn:cm_argmax}
% \small
% e_{sub}= \argmax_{\tilde{e} \in \mathbb{E}^{\ell_\text{tgt}, y}}~\textsc{Cosine}(\mathbf{Emb}(e), \mathbf{Emb}(\tilde{e}))
% \end{equation}
To generate more augmented data with diverse entities, we further apply MELM on the gold and code-mixed data. Since the training data now contains entities from multiple languages, we also prepend a language marker to the entity token to help MELM differentiate different languages, as shown in Figure~\ref{fig:MELM_multi}. 

% The original training set $\mathbb{D}_{\text{gold}}$, code-mixed dataset $\mathbb{D}_{\text{CM}}$ and augmented data generated from gold data and code-mixed data $\{ \mathbb{D}_{\text{gold}}^{\text{aug}}, \mathbb{D}_{\text{CM}}^{\text{aug}}\}$ are combined for training the NER model\lx{Is it necessary to give such details?}.

\section{Experiments}
%\lx{In this section, we }
% To evaluate the effectiveness of MELM on low-resource NER, we conduct experiments under three different settings, namely (1) monolingual NER, where the original training set contains monolingual data and the NER model is evaluated on the same language; (2) cross-lingual NER, where training data is only available in one source language and the NER model is evaluated on multiple target languages; (3) multilingual NER, where training data from multiple languages is available and the NER model is evaluated on each of the languages.

% Under monolingual settings, we finetune MLM on monolingual training data to generate augmented data, and show their effectiveness on monolingual NER. 
% Similarly, we experiment on zero-shot cross-lingual NER, where the NER model is trained on source language gold and augmented data, and evaluated on the target languages. 
% For multilingual NER, training data from multiple languages are available and the NER model is evaluated on test data from all the languages as well. In this case, we first leverage the benefits of code-mixed data by applying the proposed phrase similarity search algorithm. Subsequently, multilingual gold data and code-mixed data are used to finetune the MLM, such that more diversified augmented data can be generated.

To comprehensively evaluate the effectiveness of the proposed MELM on low-resource NER, we consider three evaluation scenarios: \textbf{monolingual}, \textbf{zero-shot cross-lingual} and \textbf{multilingual} low-resource NER. 

\subsection{Dataset}
\label{ssec:dataset}

We conduct experiments on CoNLL NER dataset~\citep{tjong-kim-sang-2002-introduction,sang2003introduction} of four languages where $\mathbb{L}$ = \{English (En), German (De), Spanish (Es), Dutch (Nl)\}. 
For each language $\ell \in \mathbb{L}$, we first sample $N$ sentences from the full training set as $\mathbb{D}^{\ell, N}_{\text{train}}$, where $N \in \{100, 200, 400, 800\}$ to simulate different low-resource levels. 
For a realistic data split ratio, we also downscale the full development set to $N$ samples as $\mathbb{D}^{\ell, N}_{\text{dev}}$.
The full test set for each language is adopted as $\mathbb{D}^{\ell}_{\text{test}}$ for evaluation.

For \textbf{monolingual} experiments on language $\ell$ with low-resource level $N \in \{100, 200, 400, 800\}$, we use $\mathbb{D}^{\ell, N}_{\text{train}}$ as the gold training data, $\mathbb{D}^{\ell, N}_{\text{dev}}$ as the development set and $\mathbb{D}^{\ell}_{\text{test}}$ as the test set. 
For \textbf{zero-shot cross-lingual} experiments with low-resource level $N \in \{100, 200, 400, 800\}$, we use $\mathbb{D}^{\text{En}, N}_{\text{train}}$ as the source language gold training data, $\mathbb{D}^{\text{En}, N}_{\text{dev}}$ as the development set and $\mathbb{D}^{\text{De}}_{\text{test}}$, $\mathbb{D}^{\text{Es}}_{\text{test}}$ and $\mathbb{D}^{\text{Nl}}_{\text{test}}$ as target language test sets. 
Under \textbf{multilingual} settings where $N$ training data from each language is available $(N \in \{100, 200, 400\})$,
we use $\bigcup_{\ell \in \mathbb{L}} \mathbb{D}^{\ell, N}_{\text{train}}$ as the gold training data, $\bigcup_{\ell \in \mathbb{L}} \mathbb{D}^{\ell, N}_{\text{dev}}$ as the development set 
and evaluate on $\mathbb{D}^{\text{En}}_{\text{test}}$, $\mathbb{D}^{\text{De},}_{\text{test}}$, $\mathbb{D}^{\text{Es}}_{\text{test}}$ and $\mathbb{D}^{\text{Nl}}_{\text{test}}$, respectively.

\subsection{Experimental Setting}
\label{ssec:exp_setting}
% \lx{\paragraph{Dataset} We conduct experiments on the widely-used CoNLL NER dataset~\citep{tjong-kim-sang-2002-introduction,sang2003introduction}. To evaluate the effectiveness of the proposed MELM on low-resource NER, we consider three evaluation scenarios: \textbf{Monolingual}, \textbf{Zero-shot Cross-lingual} and \textbf{Multilingual}. DETAILS HERE.}

\paragraph{MELM Fine-tuning}
We use XLM-RoBERTa-base~\citep{conneau2020unsupervised} with a language-modeling head to initialize MELM parameters. 
MELM is fine-tuned for 20 epochs using Adam optimizer~\citep{DBLP:journals/corr/KingmaB14} with batch size set to 30 and learning rate set to $1e-5$.

\paragraph{NER Model} 
We use XLM-RoBERTa-Large~\citep{conneau2020unsupervised} with CRF head~\citep{lample2016neural} as the NER model for our experiments\footnote{\url{https://github.com/allanj/pytorch_neural_crf}}. 
We adopt Adamw optimizer~\citep{loshchilov2018decoupled} with learning rate set to $2e-5$ and set batch size to 16. 
The NER model is trained for 10 epochs and the best model is selected according to dev set performance. 
The trained model is evaluated on test sets and we report the averaged Micro-F1 scores over 3 runs.

\paragraph{Hyperparameter Tuning}
% Our proposed method introduces 3 hyperparameters: $\eta$ controls the masking rate in the fine-tuning phase; $\mu$ controls the masking rate in the generation phase; and $R$ is the number of rounds for data augmentation. 
% We tune the hyperparameters on the dev set with grid search and set $\eta=0.7$, $\mu=0.5$ and $R=3$.

% We first tune $\eta$ and $\xi$ with grid search. We generate one round of augmented data and train a NER tagger using only the augmented data. The optimal hyperparameter setting is selected based on the F1 score on the dev set. We set $\eta=0.7$ and $\xi=0.5$. 
% Then we fix the values of $\eta$ and $\xi$ to tune $R$. We merge the augmented data generated from $R$ rounds with gold data to train a NER tagger and select the best $R$ value according to the F1 score on the dev set. $R$ is set to 3. 

% Details on the hyperparameter tuning procedure can be found in Appendix~\ref{ssec:hyper-tuning}. 

The masking rate $\eta$ in MELM fine-tuning, the Gaussian mean $\mu$ for MELM generation and the number of MELM augmentation rounds $R$ are set as 0.7, 0.5 and 3, respectively. All of these hyperparameters are tuned on the dev set with grid search. Details of the hyperparameter tuning can be found in Appendix~\ref{ssec:hyper-tuning}

\subsection{Baseline Methods}
\label{ssec:baseline}

%To elaborate the effectiveness of the proposed framework, MELM is compared with the following methods: \bing{\cchar{proposed framework 就是 MELM，能改的更简洁一些？比如To elaborate the effectiveness of MELM, we compare it with the following methods.}}
To elaborate the effectiveness of the proposed MELM, we compare it with the following methods:
\vspace{0.5mm}

%\paragraph{Gold Only}
\noindent\textbf{Gold-Only} The NER model is trained on only the original gold training set.

%\paragraph{Label-wise Substitution}
\noindent\textbf{Label-wise Substitution}~\citet{dai2020analysis} randomly substituted named entities with existing entities of the same entity type from the original training set.

%\paragraph{Entity Replacement}
\noindent\textbf{MLM-Entity} 
%Similar to~\citet{bari-etal-2021-uxla}, 
We randomly mask entity tokens and directly utilize a pretrained MLM for data augmentation without fine-tuning and labeled sequence linearization as used in MELM. The prediction of a masked entity token does not consider label information but solely relies on the context words.

%\paragraph{Context Replacement}
% \noindent\textbf{MLM-Context} Similar to~\citet{li2020conditional}, we use a pretrained MLM to predict the masked context tokens in the training samples. The entities and their labels remain unchanged.

\noindent\textbf{DAGA}~\citet{dingdaga} firstly linearized NER labels into the input sentences and then use them to train an autoregressive language model. The language model was used to synthesize augmented data from scratch, where both context and entities are generated simultaneously.

\noindent\textbf{MulDA}~\citet{liu2021mulda} fine-tuned mBART\citep{liu2020multilingual} on linearized multilingual NER data to generate augmented data with new context and entities. 
%This method is similar to~\citet{dingdaga}, but is applicable to multilingual data\lx{Is this sentence necessary?}.

\begin{table*}[t!]
\centering
\resizebox{1.8\columnwidth}{!}{
\begin{tabular}{clccccl@{\hskip 1cm}cccl}
\hline\hline
\multirow{2}{*}{\textbf{\#Gold}}   & \multirow{2}{*}{\textbf{Method}} & \multicolumn{5}{c}{\texttt{Monolingual}} & \multicolumn{4}{c}{\texttt{Cross-lingual}} \\
& & \textbf{En} & \textbf{De} & \textbf{Es} & \textbf{Nl} & \textbf{Avg}  & \textbf{En}$\rightarrow$\textbf{De} & \textbf{En}$\rightarrow$\textbf{Es} & \textbf{En}$\rightarrow$\textbf{Nl} & \textbf{Avg}\\ 
\hline
\multirow{6}{*}{100}    &Gold-Only          &50.57	&39.47	&42.93	&21.63	&38.65  &39.54	&37.40	&39.27	&38.74\\
                        &Label-wise         &61.34	&55.00	&59.54	&27.85	&50.93  &45.85	&43.74	&50.51	&46.70\\
                        &MLM-Entity         &61.22	&50.96	&61.29	&46.59	&55.02  &47.96	&45.42	&49.34	&47.57\\
                        %&MLM-Context        &59.37	&53.46	&60.42	&47.89	&55.29\\
                        &DAGA               &68.06	&59.15	&69.33	&45.64	&60.54  &52.95	&46.72	&54.63	&51.43\\
                        &MELM \textit{w/o linearize}  &70.01	&61.92	&65.07	&59.76	&64.19  &48.70	&49.10	&53.37	&50.39\\
                        &MELM \textit{(Ours)} &\textbf{75.21} &\textbf{64.12} &\textbf{75.85} &\textbf{66.57} &\textbf{70.44}   &\textbf{56.56} &\textbf{53.83} &\textbf{60.62} &\textbf{57.00}\\%~\textcolor{red}{(+15.15)}\\
                        %&                   &\textcolor{red}{(+13.87)}  &\textcolor{red}{(+9.12)}  &\textcolor{red}{(+14.56)}  &\textcolor{red}{(+18.68)}  &\textcolor{red}{(+15.15)}\\
\hline
\multirow{6}{*}{200}    &Gold-Only          &74.64	&62.85	&72.64 	&55.96	&66.52  &54.95	&51.26	&60.71	&55.64\\
                        &Label-wise         &76.82	&67.31	&78.34	&66.52	&72.25  &55.01	&53.14	&63.30	&57.15\\
                        &MLM-Entity         &79.16	&70.01	&78.45	&66.69	&73.58  &60.44	&57.72	&68.37	&62.18\\
                        %&MLM-Context        &78.56	&70.39	&78.82	&68.47	&74.06\\
                        &DAGA               &79.11	&69.82	&78.95	&68.53	&74.10  &59.58	&57.68	&65.74	&61.00\\
                        &MELM \textit{w/o linearize}    &81.77	&71.41	&80.43	&72.92	&76.63  &62.57	&63.49	&70.18	&65.41\\
                        &MELM \textit{(Ours)} &\textbf{82.91} &\textbf{72.71} &\textbf{80.46} &\textbf{77.02} &\textbf{78.27}   &\textbf{65.01} &\textbf{63.71} &\textbf{70.37} &\textbf{66.36}\\%~\textcolor{red}{(+4.21)}\\
                        %&                   &\textcolor{red}{(+3.76)}  &\textcolor{red}{(+2.32)}  &\textcolor{red}{(+1.64)}  &\textcolor{red}{(+8.55)}  &\textcolor{red}{(+4.21)}\\
\hline
\multirow{6}{*}{400}    &Gold-Only          &81.85	&70.77	&80.02 	&74.60	&76.81  &65.76	&61.57	&71.04	&66.12\\
                        &Label-wise         &84.62	&74.33	&81.01	&77.87	&79.46  &66.18	&67.43	&71.93	&68.51\\
                        &MLM-Entity         &83.82	&74.66	&81.08	&77.90	&79.37  &67.41	&70.28	&74.31	&70.67\\
                        %&MLM-Context        &84.31	&73.20	&81.20	&77.85	&79.14\\
                        &DAGA               &84.36	&72.95	&82.83	&78.99	&79.78  &66.77	&67.13	&72.40	&68.77\\
                        &MELM \textit{w/o linearize}    &85.16	&75.42	&82.34	&79.34	&80.56  &68.02	&66.01	&72.98	&69.00\\
                        &MELM \textit{(Ours)} &\textbf{85.73} &\textbf{77.50} &\textbf{83.31} &\textbf{80.92} &\textbf{81.87}   &\textbf{68.08} &\textbf{70.37} &\textbf{75.78} &\textbf{71.74}\\%~\textcolor{red}{(+2.41)}\\
                        %&                   &\textcolor{red}{(+1.11)}  &\textcolor{red}{(+2.84)}  &\textcolor{red}{(+2.11)}  &\textcolor{red}{(+3.02)}  &\textcolor{red}{(+2.41)}\\
\hline
\multirow{6}{*}{800}    &Gold-Only          &86.35	&78.35	&83.23 	&83.86	&82.95  &65.31	&68.28	&72.07	&68.55\\
                        &Label-wise         &86.72	&78.21	&84.42	&84.26	&83.40  &65.60	&72.22	&74.77	&70.86\\
                        &MLM-Entity         &86.50	&78.30	&84.09	&83.93	&83.20  &65.42	&69.10	&74.85	&69.79\\
                        %&MLM-Context        &86.91	&77.66	&83.60	&83.97	&83.04\\
                        &DAGA               &86.61	&77.66	&84.64	&84.90	&83.45  &\textbf{68.76}	&70.97	&75.02	&71.58\\
                        &MELM \textit{w/o linearize}    &87.35	&78.58	&84.59	&84.94	&83.99  &67.37	&71.53	&75.20	&71.37\\
                        &MELM \textit{(Ours)} &\textbf{87.59} &\textbf{79.32} &\textbf{85.40} &\textbf{85.17} &\textbf{84.37}   &67.95 &\textbf{75.72} &\textbf{75.25} &\textbf{72.97}\\%~\textcolor{red}{(+0.97)}\\
                        %&                   &\textcolor{red}{(+0.68)}  &\textcolor{red}{(+0.97)}  &\textcolor{red}{(+0.98)}  &\textcolor{red}{(+0.91)}  &\textcolor{red}{(+0.97)}\\
\hline\hline
\end{tabular}
}
\caption{
%Results of monolingual and zero-shot cross-lingual low-resource NER. \textbf{Avg} is the averaged result over all languages / transfer-pairs. English is the source language for cross-lingual settings.
Left side of table shows the results of monolingual low-resource NER. Right side of table shows the results of cross-lingual low-resource NER with English as source language. \textbf{Avg}s on left side and right side are the averaged result over all languages and all transfer pairs, respectively. 
}
\label{tab:mono_cross}
\end{table*}

\subsection{Experimental Results}

\subsubsection{Monolingual and Cross-lingual NER}
\label{sssec:mono}

As illustrated on the left side of Table~\ref{tab:mono_cross}, the proposed MELM consistently achieves the best averaged results across different low-resource levels, demonstrating its effectiveness on monolingual NER. Compared to the best-performing baselines, our MELM obtains 6.3, 1.6, 1.3, 0.38 absolute gains on 100, 200, 400 and 800 levels, respectively. Cross-lingual NER results are shown on the right side of Table 2. Again, on each of the designed low-resource levels, our MELM is superior to baseline methods in terms of the averaged F1 scores. We also notice that, given 100 Nl training samples, the Gold-Only method without data augmentation almost fails to converge while the monolingual F1 of our MELM reaches 66.6, suggesting that data augmentation is crucial for NER when the annotated training data is extremely scarce.

To assess the efficacy of the proposed labeled sequence linearization (Section~\ref{ssec:linearize}), we directly fine-tune MELM on masked sentences without linearization (as shown in Figure~\ref{fig:MELM_a}), denoted as MELM \textit{w/o linearize} in Table~\ref{tab:mono_cross}. We observe a considerable performance drop compared with MELM, which proves the label information injected via linearization indeed helps MELM differentiate different entity types, and generate entities compatible with the original label.
%\lx{Cross-lingual NER results appear in the right side of Table 2, Again, on each of the designed low-resource levels, our MELM is superior to all of the compared methods in terms of the averaged F1 scores.}

Taking a closer look at the baseline methods, we notice that the monolingual performance of Label-wise is still unsatisfactory in most cases. One probable reason is that only existing entities within the training data are used for replacement and the entity diversity after augmentation is not increased. Moreover, randomly sampling an entity of the same type for replacement is likely to cause incompatibility between the context and the entity, yielding a noisy augmented sample for NER training. Although MLM-Entity tries to mitigate these two issues by employing a pretrained MLM to generate novel tokens that fit into the context, the generated tokens might not be consistent with the original labels. Our MELM also promotes the entity diversity of augmented data by exploiting pretrained model for data augmentation. 

In the meantime, equipped with the labeled sequence linearization strategy, MELM augmentation is explicitly guided by the label information and the token-label misalignment is largely alleviated, leading to superior results in comparison to Lable-wise and MLM-Entity.

% \zr{We also compare with DAGA~\citep{dingdaga}, which generates augmented data using an autoregressive language model trained on gold NER data. Despite achieving some improvement, its performance is still inadequate. By inspecting the augmented data, we find the generated sentences are less fluent and grammatical, because the autoregressive language model cannot be fully trained given only a few hundreds of training samples. In contrast, MELM focuses on modifying entity tokens and leave the context unchanged, which ensures the overall readability of augmented sentences even under extremely low-resource settings.}

We also compare with DAGA~\citep{dingdaga}, which generates augmented data from scratch using an autoregressive language model trained on gold NER data. Although DAGA is competitive on low-resource levels of 400 and 800, it still underperforms the proposed MELM by a large margin when the training size reduces to 100 or 200.  We attribute this to the disfluent and ungrammatical sentences generated from the undertrained language model. Instead of generating augmented data from scratch, MELM focuses on modifying entity tokens and leave the context unchanged, which guarantees the quality of augmented sentences even under extremely low-resource settings.

\subsubsection{Multilingual NER}

% \paragraph{Dataset}
% When training on data from multiple languages concurrently, a multilingual NER model can be applied on multiple languages without having to train a separate model for each language. Under multilingual settings, we reuse the low-resource datasets in Section~\ref{sssec:mono} and combine training sets of English, German, Spanish and Dutch as our low-resource multilingual datasets. We evaluate on three low-resource levels, with the training set containing 100, 200 and 400 sentences from each language respectively. The development sets from each language are merged as well, and the original test sets are adopted for evaluation.

% Given the gold training data $\mathbb{D}_{\text{gold}}$, we first apply the proposed phrase similarity search algorithm to generate code-mixed data $\mathbb{D}_{\text{CM}}$. Then, we fine-tune MELM on ${\mathbb{D}_{\text{gold}}, \mathbb{D}_{\text{CM}}}$ to generate augmented data ${\mathbb{D}_{\text{gold}}^{\text{aug}}, \mathbb{D}_{\text{CM}}^{\text{aug}}}$. Finally, all datasets ${\mathbb{D}_{\text{gold}}, \mathbb{D}_{\text{CM}}, \mathbb{D}_{\text{gold}}^{\text{aug}}, \mathbb{D}_{\text{CM}}^{\text{aug}}}$ are combined for training the NER model. 

\begin{table}[t!]
\centering
\resizebox{0.99\columnwidth}{!}{
\begin{tabular}{clccccc}
\hline\hline
\textbf{\#Gold}   & \textbf{Method} & \textbf{En} & \textbf{De} & \textbf{Es} & \textbf{Nl} & \textbf{Avg}  \\ 
\hline
\multirow{6}{*}{100 $\times$4}    &Gold-Only          &75.62	&69.35	&75.85	&74.33	&73.79\\
                        &MulDA              &73.67	&70.47	&75.53	&72.40	&73.02\\
                        &MELM-\textit{gold (Ours)}  &78.71	&74.79	&81.25	&78.85	&78.40\\
                        &Code-Mix-\textit{random}        &77.38	&70.58	&78.61	&76.45	&75.75\\
                        &Code-Mix-\textit{ess (Ours)}             &79.55	&71.56	&79.58	&76.49	&76.80\\
                        &MELM \textit{(Ours)}        &\textbf{80.96} &\textbf{75.61} &\textbf{81.47} &\textbf{80.14} &\textbf{79.54}\\

\hline
\multirow{6}{*}{200 $\times$4}    &Gold-Only          &83.06	&76.39	&82.71	&79.19	&80.34\\
                        &MulDA              &82.32	&74.57	&82.73	&79.06	&79.67\\
                        &MELM-\textit{gold (Ours)}  &82.90	&78.05	&\textbf{85.93}	&81.00	&81.97\\
                        &Code-Mix-\textit{random}        &82.86	&75.70	&83.13	&79.08	&80.19\\
                        &Code-Mix-\textit{ess (Ours)}            &83.34	&76.64	&82.02	&82.27	&81.07\\
                        &MELM \textit{(Ours)}        &\textbf{83.56} &\textbf{78.24} &84.98 &\textbf{82.79} &\textbf{82.39}\\

\hline
\multirow{6}{*}{400 $\times$4}    &Gold-Only          &83.92	&77.40	&83.22	&84.04	&82.14\\
                        &MulDA              &84.37	&78.41	&84.54	&83.09	&82.60\\
                        &MELM-\textit{gold (Ours)}  &86.04	&79.09	&85.76	&84.83	&83.93\\
                        &Code-Mix-\textit{random}        &85.04	&77.91	&84.44	&83.56	&82.74\\
                        &Code-Mix-\textit{ess (Ours)}            &85.74	&80.03	&85.18	&85.36	&84.08\\
                        &MELM \textit{(Ours)}        &\textbf{86.14} &\textbf{80.33} &\textbf{86.60} &\textbf{85.99} &\textbf{84.76}\\

\hline\hline
\end{tabular}
}
\caption{Results of multilingual low-resource NER. Gold training set contains the same number of training samples from each language. \textbf{Avg} is the averaged result over all languages.}
\label{tab:multi}
\end{table}

For multilingual low-resource NER, we firstly directly apply MELM on the concatenation of training sets from multiple languages. As shown in Table~\ref{tab:multi}, MELM-\textit{gold} achieves substantial improvement over the Gold-only baseline, which is consistent with monolingual and cross-lingual results. We compare with MulDA~\citep{liu2021mulda} as a baseline data augmentation method. MulDA generates augmented data autoregressively with an mBART model, which is fine-tuned on NER data with inserted label tokens. At the low-resource levels in our experimental settings, MulDA is less effective and even leads to deteriorated performance. The unsatisfactory performance mainly results from the discrepancy between pretraining and fine-tuning due to the inserted label tokens. Given very few training samples, it is difficult to adapt mBART to capture the distribution of the inserted label tokens, and thus MulDA struggles to generate fluent and grammatical sentences from scratch. In comparison, our proposed method preserves the original context and introduce less syntactic noise in the augmented data.
To further leverage the benefits of code-mixing in multilingual NER, we experiment with two code-mixing methods: (1) Code-Mix-\textit{random}, which randomly substitutes entities with existing entities of the same type from other languages, and (2) Code-Mix-\textit{ess}, which adopts the proposed entity similarity search algorithm in Section~\ref{ssec:code-mix} as the code-mixing strategy. 

Experimental results in Table~\ref{tab:multi} show that both methods are able to achieve improved performance over Gold-Only. This observation suggests that code-mixing techniques, either random code-mixing or code-mixing via our entity similarity search, are indeed helpful for multilingual NER. Comparing these two methods, the performance gains brought by Code-Mix-\textit{ess} are more significant and consistent across different low-resource levels, which demonstrates the effectiveness of our proposed entity similarity search algorithm. Applying MELM on both gold data and code-mixed data from Code-Mix-\textit{ess}, the multilingual NER results are further improved. In summary, our proposed MELM is well-suited for multilingual NER, which can be integrated with our code-mixing technique to achieve further improvement.

\section{Further Analysis}
\label{sec: analysis}

\subsection{Case Study}
\label{ssec: analysis_align}

\begin{table*}[t!]
\centering
\resizebox{2\columnwidth}{!}{
\begin{tabular}{cllllll} 
\hline\hline
\textbf{\textcolor{blue}{Text}} & \textcolor{blue}{EU} & \textcolor{blue}{rejects} & \textcolor{blue}{German} & \textcolor{blue}{call to boycott}  & \textcolor{blue}{British} & \textcolor{blue}{Lamb}\\
\textbf{\textcolor[rgb]{1,0.643,0}{Label}} & \textcolor[rgb]{1,0.643,0}{B-ORG} & \textcolor[rgb]{1,0.643,0}{O} & \textcolor[rgb]{1,0.643,0}{B-MISC} & \textcolor[rgb]{1,0.643,0}{O~~~O~~~O} & \textcolor[rgb]{1,0.643,0}{B-MISC} & \textcolor[rgb]{1,0.643,0}{O}\\ 
\hline
\textbf{MLM} & \multicolumn{2}{l}{\makecell[l]{\textcolor{red}{Britain,~}EU,\textcolor{red}{UK,~} \textcolor{red}{Trump,~}\textcolor{red}{US}}} &  \multicolumn{2}{l}{\makecell[l]{\textcolor{red}{US,~}\textcolor{red}{a,~}\textcolor{red}{UN,~}  \textcolor{red}{the,~}\textcolor{red}{UK}}}  &  \multicolumn{2}{l}{\makecell[l]{\textcolor{red}{the,~}\textcolor{red}{a,~}\textcolor{red}{black,~} \textcolor{red}{white,~}\textcolor{red}{young}}}\\ 
\hline
\makecell[c]{\textbf{MELM} \\ \textit{w/o linearize}} & \multicolumn{2}{l}{\makecell[l]{EU, \textcolor{red}{Australia,~}\textcolor{red}{US,~} UN, \textcolor{red}{Israel}}} &  \multicolumn{2}{l}{\makecell[l]{German, Indian, \textcolor{red}{the,~}  \textcolor{red}{Washington,~}\textcolor{red}{Union}}}  &  \multicolumn{2}{l}{\makecell[l]{Chinese, British, \textcolor{red}{raw,~} \textcolor{red}{California,~}Australian}}\\ 
\hline
\textbf{MELM} & \multicolumn{2}{l}{\makecell[l]{EU, Greenpeace, Amnesty, UN, Reuters}} & \multicolumn{2}{l}{\makecell[l]{German, British, Dutch, French, \textcolor{red}{EU}}} & \multicolumn{2}{l}{\makecell[l]{African, British, Guinean, \textcolor{red}{white,~} French}} \\ 
\hline\hline
\textbf{\textcolor{blue}{Text}} & \textcolor{blue}{Clinton} & \textcolor{blue}{aide} & \textcolor{blue}{resigns~,} & \textcolor{blue}{NBC} & \textcolor{blue}{says} \\
\textbf{\textcolor[rgb]{1,0.643,0}{Label}} & \textcolor[rgb]{1,0.643,0}{B-PER} & \textcolor[rgb]{1,0.643,0}{O} & \textcolor[rgb]{1,0.643,0}{O~~~~~~~~O}  & \textcolor[rgb]{1,0.643,0}{B-ORG} & \textcolor[rgb]{1,0.643,0}{O}\\ 
\hline
\textbf{MLM} & \multicolumn{3}{l}{\makecell[l]{\textcolor{red}{my,~his,~My,} \textcolor{red}{When,~her}}} & \multicolumn{3}{l}{\makecell[l]{\textcolor{red}{he,~she,~it,} \textcolor{red}{and,~who}}} \\
\hline
\makecell[c]{\textbf{MELM} \\ \textit{w/o linearize}}  & \multicolumn{3}{l}{\makecell[l]{\textcolor{red}{French,~German,~British,~} \textcolor{red}{Swiss,~Russian}}} & \multicolumn{3}{l}{\makecell[l]{Reuters, \textcolor{red}{Pompeo,~Blair~} \textcolor{red}{Hill,~} AFP}} \\
\hline
\textbf{MELM} & \multicolumn{3}{l}{\makecell[l]{\textcolor{red}{French,~} White, Walker, Ferguson, David}} & \multicolumn{3}{l}{\makecell[l]{NBC, AFP, Greenpeace, BBC, Anonymous}} \\
\hline\hline
\end{tabular}}
\caption{Examples of the top-5 predictions by MLM, MELM \textit{w/o linearize} and MELM. Predictions that do not belong to the original class are highlighed in red.}
\label{tab:case_study}
\end{table*}

Apart from the quantitative results, we further analyze the augmented data to demonstrate the effectiveness of our MELM in maintaining the consistency between the original label and the augmented token. Table~\ref{tab:case_study} presents examples of the top-5 predictions from pretrained MLM, MELM \textit{w/o linearize} and MELM. As we can see, the pretrained MLM, which does not introduce any design or contraint on data augmentation, tends to generate high-frequency words such as ``the'', ``he'' and ``she'', and the majority of generated words do not belong to the original entity class. Being finetuned on NER data with entity-oriented masking, MELM \textit{w/o linearize} is able to generate more entity-related tokens. 

However, without the explicit guidance from entity labels, it is still too difficult for MELM \textit{w/o linearize} to make valid predictions solely based on the ambiguous context (e.g., both ``Pompeo'' (PER) and ``Reuters'' (ORG) are compatible with the context of Example \#2), which leads to token-label misalignment. Compared to the above methods, our MELM take both label information and context into consideration, and thus generates more entities that fit into the context and align with the original label as well. Moreover, it is noteworthy that MELM can leverage the knowledge from pretrained model to generate real-world entities that do not exist in the original NER dataset (e.g., ``Greenpeace'' and ``Amnesty''), which essentially increases the entity diversity in training data.

\subsection{Number of Unique Entities}

\begin{figure}[t]
    \centering
    \includegraphics[width=0.9\columnwidth]{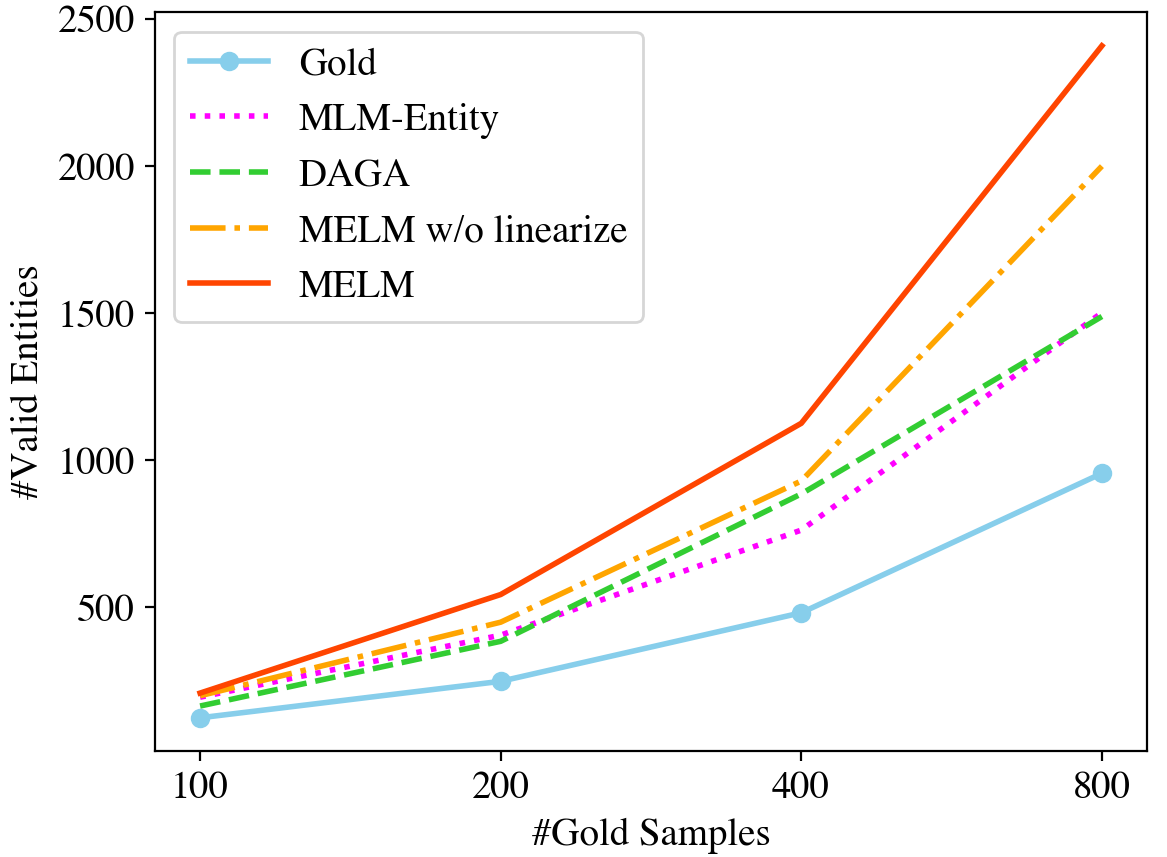}
    \caption{Comparison between the number of unique valid entities introduced by different methods}
    \label{fig:new_entity}
\end{figure}

As demonstrated in~\citet{lin2020rigorous} and our preliminary experiments in Figure~\ref{fig:evc}, introducing unseen entities can effectively provide more entity regularity knowledge, and helps to improve NER performance. Therefore, we examine the amount of unique entities introduced by different methods. As there might be token-label misalignment in the augmented data, we firstly train an `oracle' NER model on the full CoNLL dataset and then use it to tag training data of MELM and different baseline methods. For each method, we count the total number of unique entities whose labels match the labels assigned by the `oracle' model. As shown in Figure~\ref{fig:new_entity}, while many augmented entities from MLM-Entity, DAGA and MELM \textit{w/o linearize} are filtered out due to token-label misalignment, we note that MELM introduces a significantly larger number of unseen entities in the augmented data. Therefore MELM is able to provide richer entity regularity knowledge, which explains its superiority over the baseline methods. 

\vspace{-1mm}

$ $

\section{Related Work}

% \textbf{Word-replacement-based Data Augmentation}~\citet*{wei2019eda} applied synonym replacement based on WordNet~\citep{miller1998wordnet} to sentence classification tasks, and achieved comparable results with only 50\% full dataset. However, the diversity of synonym replacement is limited by the knowledge base.~\citet{kobayashi2018contextual} generated augmented data for sentence classification tasks, by replacing words with paradigmatically-related words using a bi-directional LSTM language model conditioned on the sentence label.~\citet{wu2019conditional} extended language-model-based word replacement from LSTM to Transformer models (BERT) and added sentence label embedding to all tokens, to condition word replacement on the sentence label.~\citet{kumar2020data} further generalized language-model-based word replacement to both auto-encoding and auto-regressive transformer models. They prepended the sentence with sentence label during training and augmentation, to incorporate sentence class information. 

On sentence level tasks, one line of data augmentation methods are built upon word-level modifications, which can be based on synonym replacement~\citep{wei2019eda}, LSTM language model~\citep{kobayashi2018contextual}, MLM~\citep{wu2019conditional, kumar2020data}, auto-regressive pretrained LM~\citep{kumar2020data}, or constituent-based tagging schemes~\citep{zhoext}. However, these methods suffer from token-label misalignment when applied to token-level tasks such as NER, which requires sophisticated post-processing to remove noisy samples in augmented data~\citep{bari-etal-2021-uxla,xshbook}. 

Existing works avoid token-label misalignment by replacing entities with existing entities of the same class~\citep{dai2020analysis}, or only modifying context works and leaving entities / aspect terms unchanged~\cite{li2020conditional}. Others attempt to produce augmented data by training / fine-tuning a generative language model on linearized labeled sequences~\citep{dingdaga, liu2020multilingual}.

Backtranslation~\citep{sennrich2015improving,fadaee2017data,dong2017learning,yu2018fast} translates source language sentences into a target language, and subsequently back to the source language, which preserve the overall semantics of the original sentences. On token-level tasks, however, they hinge on external word alignment tools for label propagation, which are often error-prone~\citep{tsai2016cross, li2020unsupervised}.

\section{Conclusion}
We have proposed MELM as a data augmentation framework for low-resource NER. Through labeled sequence linearization, we enable MELM to explicitly condition on label information when predicting masked entity tokens. Thus, our MELM effectively alleviates the token-label misalignment issue and generates augmented data with novel entities by exploiting pretrained knowledge. Under multilingual settings, we integrate MELM with code-mixing for further performance gains. Extensive experiments show that the proposed framework demonstrates encouraging performance gains on monolingual, cross-lingual and multilingual NER across various low-resource levels.

\section*{Acknowledgements}
\label{sec:acknowledgements}

%\vspace{-0.5em}
This research is partly supported by the Alibaba-NTU Singapore Joint Research Institute, Nanyang Technological University. Erik Cambria would like to thank the support by the Agency for Science, Technology and Research (A*STAR) under its AME Programmatic Funding Scheme (Project \#A18A2b0046).

\bibliography{anthology,custom}
\bibliographystyle{acl_natbib}

\clearpage
\pagenumbering{arabic}

\appendix

\section{Appendix}
\label{sec:appendix}

\subsection{Hyperparameter Tuning}
\label{ssec:hyper-tuning}

\textbf{Masking hyperparameters.} To determine the optimal setting for fine-tune mask rate $\eta$ and generation masking parameter $\mu$, we conduct a grid search on both hyperparameters in range $[0.3, 0.5, 0.7]$. We finetune MELM and generate English augmented data on CoNLL following our method in Section~\ref{sec: method}. The augmented data is used to train a NER tagger and its performance on English dev set is recorded. As shown in Table~\ref{tab:mask_tune}, we achieve the best dev set F1 when $\eta=0.7$ and $\mu=0.5$, which is adopted for the rest of this work.

\begin{table}[h]
\centering
\begin{tabular}{|ll|lll|} 
\hline
 &  &  & \textbf{$\eta$} &  \\
 &  & 0.3 & 0.5 & 0.7 \\ 
\hline
 & 0.3 & 76.90 & 75.64 & 78.08 \\
\textbf{$\mu$} & 0.5 & 76.16 & 78.06 & \textbf{78.56} \\
 & 0.7 & 75.94 & 78.09 & 78.37 \\
\hline
\end{tabular}
\caption{Dev set F1 for masking hyperparameter tuning.}
\label{tab:mask_tune}
\end{table}

\noindent \textbf{Number of augmentation rounds.} Merging augmented data from multiple rounds increase entity diversity until it saturates at certain point. Continuing adding in more augmented data begins to amplify the noise in augmented data and leads to decreasing performance. To determine the optimum number of augmentation rounds $R$, we merge different amount of augmented data with English gold data to train a NER tagger, with $R$ ranging from 1 to 6. As shown in Table~\ref{tab:aug_round}, dev set F1 increases with increasing amount of augmented data until $R$=3, and starts to drop further beyond. Therefore, we choose $R=3$ for all of our experiments.

\begin{table}[h]
\centering
\resizebox{0.99\columnwidth}{!}{
\begin{tabular}{lllllll}
\hline
\textbf{R} & 1 & 2 & 3 & 4 & 5 & 6 \\
\hline
\textbf{Dev F1} & 92.35 & 92.36 & \textbf{92.84} & 92.72 & 92.59 & 92.39 \\
\hline
\end{tabular}}
\caption{Dev set F1 for number of augmentation rounds.}
\label{tab:aug_round}
\end{table}

\subsection{Statistics for Reproducibility}

In this section, we present the validation F1 averaged among 3 runs of MELM under different languages and low-resource levels. We also summarize the estimated time for fine-tuning MELM and the number of parameters used. We separately show the statistics of monolingual (Table~\ref{tab:mono_val}), cross-lingual (Table~\ref{tab:cross_val}) and multilingual (Table~\ref{tab:multi_val}) NER.

\begin{table}[h]
\centering
\resizebox{0.99\columnwidth}{!}{
\begin{tabular}{cccccccc}
\hline\hline
\textbf{\#Gold} & \textbf{En} & \textbf{De} & \textbf{Es} & \textbf{Nl} &\textbf{time} &\textbf{\#Paramerter}\\ 
\hline
100    &82.38	&71.11	&71.77	&71.01  &\textasciitilde{} 7min & 270M\\
200    &85.93	&77.96	&83.25	&79.53  &\textasciitilde{} 10min & 270M\\
400    &89.01	&82.95	&85.10	&81.40  &\textasciitilde{} 15min & 270M\\
800    &92.01	&84.82	&86.65	&85.61  &\textasciitilde{} 20min & 270M\\
\hline\hline
\end{tabular}
}
\caption{Validation F1 for MELM under monolingual settings}
\label{tab:mono_val}
\end{table}

\begin{table}[h]
\centering
\resizebox{0.7\columnwidth}{!}{
\begin{tabular}{cccccccc}
\hline\hline
\textbf{\#Gold} & \textbf{dev F1} &\textbf{time} &\textbf{\#Paramerter}\\ 
\hline
100    &82.38   &\textasciitilde{} 7min & 270M\\
200    &85.93   &\textasciitilde{} 10min & 270M\\
400    &89.01	&\textasciitilde{} 15min & 270M\\
800    &92.01	&\textasciitilde{} 20min & 270M\\
\hline\hline
\end{tabular}
}
\caption{Validation F1 for MELM under cross-lingual settings}
\label{tab:cross_val}
\end{table}

\begin{table}[h!]
\centering
\resizebox{0.99\columnwidth}{!}{
\begin{tabular}{cccccccc}
\hline\hline
\textbf{\#Gold per language} & \textbf{dev F1} &\textbf{time} &\textbf{\#Paramerter}\\ 
\hline
100    &83.21   &\textasciitilde{} 20min & 270M\\
200    &84.83   &\textasciitilde{} 30min & 270M\\
400    &87.07	&\textasciitilde{} 45min & 270M\\
\hline\hline
\end{tabular}
}
\caption{Validation F1 for MELM under multilingual settings}
\label{tab:multi_val}
\end{table}

\subsection{Computing Infrastructure}

Our experiments are conducted on NVIDIA V100 GPU.

\end{document}